\pdfoutput=1
\documentclass[11pt]{article}
\usepackage[final]{acl}

\usepackage{latexsym}
\usepackage[linesnumbered,ruled,plain]{algorithm2e}

\usepackage{enumitem}
\definecolor{lightorange}{RGB}{255,200,150}
\usepackage[T1]{fontenc}
\usepackage[utf8]{inputenc}
\usepackage{microtype}

\usepackage{cuted}
\usepackage{float}
\usepackage{threeparttable}
\usepackage{siunitx}
\usepackage{etoolbox}

\robustify\tnote
\usepackage{multirow}

\usepackage{amsmath}

\usepackage{kotex}
\usepackage[most]{tcolorbox}

\definecolor{highlightgreen}{HTML}{C9F3C6}

\usepackage{graphicx}
\usepackage{tabularx}
\usepackage{booktabs}

\usepackage{times}

\title{CAC-CoT: Connector-Aware Compact Chain-of-Thought for Efficient Reasoning Data Synthesis Across Dual-System Cognitive Tasks}
\author{
  \textbf{Sunguk Choi}\thanks{Equal contribution} \quad
  \textbf{Yonghoon Kwon}\footnotemark[1] \quad
  \textbf{Heondeuk Lee}\footnotemark[1] \\
  DATUMO \\
  \texttt{\{sunguk.choi, yonghoon.kwon, heondeuk.lee\}@selectstar.ai}
}

\begin{document}
\maketitle
\begin{abstract}

Long chain-of-thought (CoT) prompting helps Large Language Models (LLMs) solve difficult problems, but very long traces often slow or even degrade performance on fast, intuitive “System-1” tasks. We introduce Connector-Aware Compact CoT (CAC-CoT) --- a method that deliberately restricts reasoning to a small, fixed set of connector phrases, steering the model toward concise and well --- structured explanations. Despite its simplicity, our synthetic method with general-purpose LLMs yields a high-quality training quality. CAC-CoT achieves $\approx$ 85\% on GSM8K and $\approx$ 40\% on GPQA (System-2) while also achieving $\approx$ 85\% on S1-Bench (System-1), surpassing the baseline by over 20\%. Its reasoning traces average $\approx$ 300 tokens(ART), about one-third the length of baseline traces, delivering higher efficiency without loss of accuracy.\footnote{Code is publicly available at 
\url{https://github.com/selectstar-ai/CAC-CoT}} \end{abstract}

\section{Introduction}

\begin{figure}[t]
  \includegraphics[width=1\columnwidth]{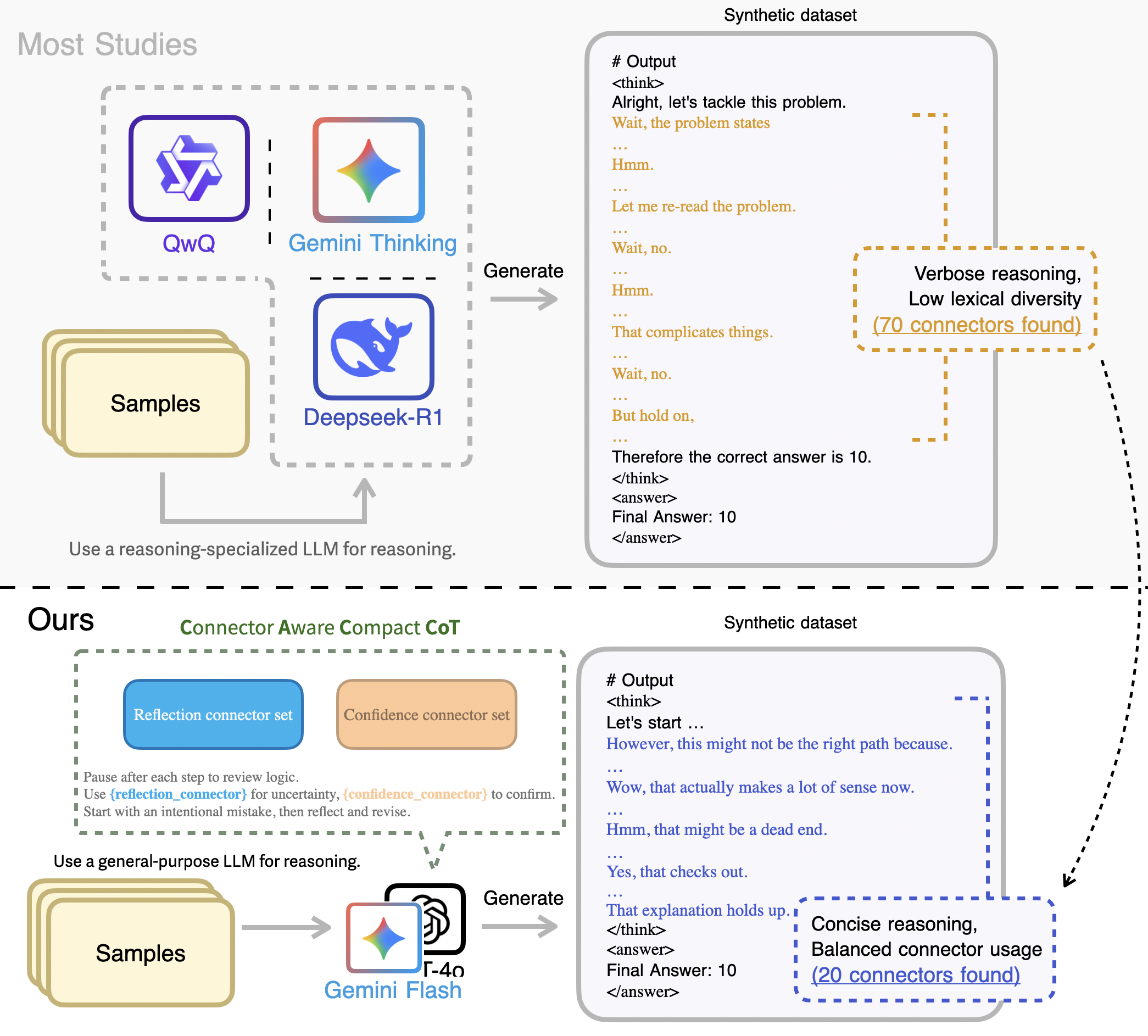}
  \caption{\textbf{Comparison of reasoning trace generation.} Most studies (top) use standard LLMs to generate verbose, repetitive reasoning with excessive connector usage. In contrast, our Connector-Aware Compact CoT framework (bottom) employs explicit connector control to produce concise, coherent reasoning traces with significantly fewer connectors, enabling efficient and high-quality data generation.}
  \label{fig:overview}
\end{figure}

Large Language Models (LLMs) have achieved striking gains on reasoning tasks by producing explicit chain-of-thoughts (CoT) rationales \cite{wei2022chain}. For complex problems that demand slow, analytical System-2 thinking, they deploy Long CoTs enriched with self-reflection, backtracking and budget-forcing --- an approach adopted by frontier systems such as OpenAI’s o1 \cite{jaech2024openai}, DeepSeek-R1 \cite{guo2025deepseek} and the inference-time scaling method s1 \cite{muennighoff2025s1} --- and one that has markedly improved performance on challenging reasoning benchmarks.

Frontier Reasoning LLMs --- e.g., from OpenAI, Anthropic and Google --- internalize long-chain reasoning techniques and dynamically adjust CoT length, thereby achieving state-of-the-art results on demanding benchmarks like AIME and GPQA \citep{jaech2024openai, anthropic2025claude37, google2025gemini25}. Recent open-source initiatives follow the same trajectory. Representative efforts include RedStar \cite{xu2025redstar} and SkyThought \cite{li2025llm}, which boost reasoning performance by fine-tuning on large Long CoT corpora with reinforcement learning (RL), while LIMR \cite{li2025limr} shows that carefully selected subsets can yield comparable gains at lower cost. Complementary work demonstrates that small and curated reasoning datasets alone can endow compact models with strong reasoning skills via supervised fine-tuning (SFT); notable examples include LIMO and the inference-time scaling method s1 \citep{ye2025limo, muennighoff2025s1}.

Despite their promise, integrating Long-CoT reasoning into LLMs introduces significant drawbacks. Models fine-tuned on very long traces tend to generate equally lengthy chains at inference time, a phenomenon termed overthinking --- they continue to reason well past the point of finding a correct answer \cite{chen2025towards}. Such gratuitous deliberation inflates computation and may even degrade accuracy, as unnecessary steps accumulate noise and contradictions. Recent evaluations corroborate this trade-off: Long-CoT-optimized models excel on System-2 thinking benchmarks yet falter on intuitive, System-1 tasks \cite{zhang2025s1bench}. These findings underscore a pressing need for balanced dual-system reasoning in LLMs --- models must learn to invoke fast, heuristic reasoning for simple queries while reserving slow, analytical reasoning for genuinely hard problems.

In this paper, we propose \textbf{Connector-Aware Compact Chain-of-Thoughts} (CAC-CoT), a data-generation framework that bridges the gap between analytical System-2 reasoning and agile System-1 intuition.  CAC-CoT prompts frontier LLMs to produce reasoning traces that (i) insert explicit connector sentences --- e.g., “Let’s pause and rethink this.”, “Wow, that actually makes a lot of sense now.” --- as deliberate checkpoints for self-reflection, whether in correct or incorrect reasoning, and (ii) enforce structural compactness that discourages gratuitous length.  These connectors cue the model to pause, verify, or backtrack only when necessary, preventing runaway verbosity while preserving logical coherence.  In effect, it trains models to balance dual-system reasoning: they engage thorough analytical reasoning when warranted yet default to concise, intuitive responses on simpler queries. The proposed method is outlined in Figure \ref{fig:overview}.

To build a high-quality CAC-CoT dataset, we leverage general-purpose LLMs --- cost-efficient but capable frontier models --- using a simple single-turn prompt that elicits concise, connector-rich traces. Despite the dataset's modest size and simplicity of the prompting strategy, fine-tuning target models on this data yields strong reasoning performance. The result is an exceptionally economical training recipe that bypasses the need for large Long CoT corpora or complex pipelines.

Comprehensive experiments confirm CAC-CoT’s effectiveness.  On demanding System-2 thinking benchmarks, our model attains $\approx$ 85\% on GSM8K and $\approx$ 40\% on GPQA \cite{rein2024gpqa} with Long-CoT fine-tuned specialists.  On S1-Bench \cite{zhang2025s1bench}, which stresses intuitive System-1 thinking, Ours reaches $\approx$ 85 \%, representing an improvement of over 20\% points compared to the s1.1 and LIMO baselines.  Efficiency follows suit: the model averages an ART of $\approx$ 300, surpassing nearly all prior systems in computational economy.  Collectively, these results demonstrate that connector-based, compact reasoning traces are a promising foundation for future reasoning data sets and models.

\section{Related works}

\subsection{Long Chain-of-Thought}
Recent advancements in CoT prompting have significantly enhanced the reasoning abilities of Large Language Models (LLMs) by decomposing complex tasks into intermediate steps \cite{wei2022chain}. Early work mainly relied on static, human-written exemplars, but subsequent studies introduced self-reflection and backtracking mechanisms to generate longer yet more adaptive reasoning traces. For instance, Reflexion prompts an agent to store episodic self-feedback between trials \citep{shinn2023reflexion}, whereas Self-Refine iteratively rewrites its own output until self-criticism converges \citep{madaan2023self}. Building on this idea, \citet{pang2024iterative} directly optimizes pairwise preferences between good and bad traces, and \citet{adarsh2024siked} distills high-quality rationales into smaller models via self-guided cycles of error detection and re-generation.

In other research directions on Long CoT, there is ongoing debate regarding the optimal length and granularity of reasoning traces. \cite{yao2023tree} argue that concise, structured traces improve efficiency without harming accuracy, whereas \cite{shen2025long} and \cite{snell2024scalingllmtesttimecompute} present scaling laws showing that longer reasoning often improves performance on difficult tasks, particularly for large models. Recent observations suggest that overly lengthy reasoning traces can result in diminishing returns or even performance degradation due to noise accumulation and increased inference costs \citep{chen2025towards}. This phenomenon, often termed overthinking, disproportionately hampers smaller models, which struggle to assimilate and faithfully reproduce the excessively long reasoning chains that arise when they are trained on Long CoT data \citep{li2025small, wu2025more, xu2025phi, zhang2025s1bench}.

\subsection{Data Efficiency and Overthinking}

While architectural advances have driven recent gains in reasoning performance, growing evidence highlights that the structure and quality of training data play an equally critical role \citep{li2025llm, swayamdipta2020dataset}. In particular, Long Chain-of-Thought (Long CoT) supervision has proven effective for inducing structured reasoning. However, such traces are often verbose, increasing both training cost and the risk of overfitting to unnecessary steps \citep{chen2025towards}.

Recent studies attempt to mitigate this verbosity by emphasizing logical sufficiency over exhaustiveness. For instance, Sky-T1 \citep{li2025llm} enhances reasoning robustness by filtering out incorrect reasoning traces rather than compressing them. Meanwhile, LIMO \citep{ye2025limo} curates high-quality datasets and applies multi-step supervision to highlight only the essential logical steps. Although neither approach directly targets brevity, both demonstrate the potential to improve training efficiency by focusing on core reasoning content. These strategies reflect a broader trend toward data-centric LLM development, where the structure and quality of supervision play a decisive role in downstream reasoning behavior.

Despite improvements, overthinking remains a failure mode. Models often produce overly detailed reasoning even for intuitive tasks, leading to redundancy, inconsistencies, and degraded performance—particularly on System-1 benchmarks \citep{sui2025stop, zhang2025s1bench}. This issue is pronounced in smaller models, where excessive elaboration introduces noise or contradictions \citep{yao2023beyond, chen2024not}. Even compact prompting strategies can backfire when misapplied to simple problems, underscoring the need for task-aware reasoning generation that adjusts explanation depth to problem complexity.

\subsection{Dual-System Reasoning}

Dual-system theory distinguishes between two types of cognition: fast, intuitive System-1 reasoning, and slow, analytical System-2 reasoning \citep{kahneman2011thinking, kannengiesser2019design}. Although large reasoning models excel at complex System-2 tasks, their performance often deteriorates on simpler System-1 tasks due to unnecessarily elaborate reasoning strategies \citep{zhang2025s1bench}. This highlights a critical limitation—current models often lack the cognitive flexibility needed to generalize across varied reasoning demands.

Early attempts to address this limitation took divergent approaches. LIMO curated a minimal yet challenging question set requiring detailed reasoning and manually verified each reasoning step for high-quality supervision. However, its scalability is limited by the labor-intensive nature of manual annotation. In contrast, Sky-T1 distilled reasoning patterns into fixed-length traces via offline analysis and trained LoRA adapters, sacrificing adaptability and stylistic diversity for inference simplicity.

CAC-CoT adopts a different approach by removing reliance on large reasoning models (LRMs) during trace generation. Instead of imitating model-generated reasoning trajectories, CAC-CoT explicitly injects connector phrases into the generation process. This strategy leverages existing high-quality question datasets (such as those from LIMO) while guiding models to produce concise, cognitively aligned reasoning traces. Notably, CAC-CoT achieves this without requiring chain-level annotations or reinforcement learning, thereby promoting both training efficiency and dual-system adaptability. Its connector-driven approach allows for controllable variation in reasoning form while maintaining semantic coherence, offering a scalable path toward more flexible and cognitively grounded reasoning systems.

\section{Methodology}

Building on recent advances in reasoning models --- which demonstrate significant performance gains through mechanisms such as self-reflection, backtracking, and self-correction --- ours, \textbf{CAC-CoT}, adopts a structured approach that explicitly injects diverse connector phrases into the training data. By deliberately guiding the model’s reasoning behavior through these connectors and tightly controlling the length of reasoning traces compared to conventional Long-CoT datasets, CAC-CoT enables the emergence of a robust dual-system reasoning capability. This design allows the model to adaptively balance concise, intuitive responses for System-1 tasks with deeper, structured reasoning for System-2 challenges, thereby achieving strong and generalizable performance across both cognitive regimes. 
The synthetic data generation prompts and the corresponding generation logic can be found in Appendix~\ref{appendix-b}, in Table~\ref{tab:prompt_table} and Algorithm~\ref{alg:synthetic_algo}, respectively.

\subsection{Connector-Aware CoT}
In reasoning models, a variety of connectors (e.g., Wait, Hmm, Alternatively) are often used to facilitate the flow of reasoning, particularly through mechanisms such as self-reflection and backtracking. Inspired by this observation, we propose a data construction method that explicitly injects connectors into the reasoning process. This encourages the model to maintain or even enhance its reasoning performance by following a structured and reflective approach. The key components of our method are as follows:
(1) Insert checkpoints after each reasoning step for reassessment;
(2) use reflection connectors at checkpoints to signal uncertainty and enable revision of faulty logic;
(3) use confidence connectors when prior logic is valid to confirm reasoning and move forward;
(4) begin with an intentionally flawed reasoning path to promote reflective correction and generate extended traces.

By prompting verification at every reasoning step and initially inducing errors to ensure that reasoning chains remain appropriately maintained, and by applying confidence and reflection connectors as needed, we steer the process to achieve both exploration and convergence. 
Details on connector usage are provided in Section4.4, and the types of connectors are summarized in Table~\ref{tab:connectors} of Appendix~\ref{appendix-b}.

\subsection{Compact CoT}
Reasoning models often suffer from excessively long reasoning traces, which can lead to inefficiency and even performance degradation on System-1 thinking tasks. To mitigate this issue, impose explicit termination constraints by limiting the number of validations, bounding the trace length, and defining clear stopping conditions. Moreover, the two types of connectors --- reflection connector and confidence connector --- further enhance the compactness of the reasoning steps, with the confidence connector in particular facilitating concise reasoning process.

By alternating these two types of connectors, selectively guide the model to either expand or converge its reasoning process. This prevents unnecessary elaboration and encourages timely termination, preserving overall reasoning performance while significantly improving performance on System-1 tasks. The termination strategy consists of the following rules:
(1) Disallow consecutive use of connectors to avoid incoherent chaining of reasoning steps;
(2) skip further validation if the same answer is produced more than once;
(3) invoke the termination condition if the reasoning becomes unclear or overly repetitive; and
(4) trigger termination if the reasoning trace exceeds a predefined length or the number of validations surpasses a threshold.

By preventing reasoning steps from expanding excessively, we ensure suitably concise reasoning traces. In particular, by imposing length limits and instructions to avoid overly repetitive steps, we achieve efficient reasoning. While triggering the termination condition does not directly improve accuracy, it prevents trace length blow-up during data generation. As shown in Section 4.4, the confidence connector terminates further progression upon successful inference, thereby promoting compactness.

\section{Experiments}

\begin{table*}[t]
\centering
\caption{
\textbf{System 1 thinking performance across models on S1-Bench.}
Comparison of evaluation metrics (Acc@5, Pass@1, Success, ART) for English (EN) and Chinese (ZN) tasks. Qwen-2.5-7B-Instruct is evaluated under loose formatting, while all other models are evaluated under strict formatting. CAC-CoT-7B (Ours) achieves strong accuracy with the lowest average reasoning length (ART), demonstrating superior efficiency and balanced performance across all categories.
\textbf{Formatting:} The AVG row is highlighted in \textbf{bold}, while notable values in all other rows are \underline{underlined} for emphasis.
}
\resizebox{\textwidth}{!}{
    \begin{tabular}{c|c|cccc|cccc}   % ← l → c 로 변경
    \specialrule{1.5pt}{0pt}{0pt}
      \addlinespace[2pt]
    \multirow{2}{*}{\textbf{Models}} &
    \multirow{2}{*}{\textbf{Task Type}} &
    \multicolumn{4}{c|}{\textbf{EN}} &
    \multicolumn{4}{c}{\textbf{ZN}} \\
    & & Acc@5\space$\uparrow$ & Pass@1\space$\uparrow$ & Success\space$\uparrow$ & ART\space$\downarrow$
      & Acc@5\space$\uparrow$ & Pass@1\space$\uparrow$ & Success\space$\uparrow$ & ART\space$\downarrow$ \\
    
    \midrule
    \multirow{5}{*}{Qwen2.5-7B-Instruct}
    & analysis\_question        & 100.0 & 100.0 & 100.0 & 49.8  & 94.44 & 96.39 & 100.0 & 37.76 \\
    & instruction\_following    &  26.47 &  57.06 & 100.0 &  6.79 & 13.79 & 21.38 & 100.0 & 10.54 \\
    & knowledge\_question       &  62.75 &  80.00 & 100.0 & 48.40 & 13.21 & 25.28 & 100.0 & 46.02 \\
    & reasoning\_question       &  66.67 &  74.67 & 100.0 & 67.08 & 41.67 & 62.08 & 100.0 & 51.61 \\
    \cmidrule(lr){2-10}
    & AVG       & 63.97 & 77.93 & 100.0 & 43.02 & 40.78 & 51.28 & 100.0 & 36.48 \\
    
    \midrule[0.8pt]
    \multirow{5}{*}{Bespoke-Stratos-7B}
    & analysis\_question       & \underline{100.0} & \underline{100.0} & \underline{100.0} & 830.43 & 75.0 & 95.24 & 99.17 & 408.97 \\
    & instruction\_following    & 58.82 & 96.69 & 88.82 & 1026.77 & \underline{65.52} & 95.59 & 93.79 & 771.93 \\
    & knowledge\_question       & \underline{100.0} & \underline{100.0} & \underline{100.0} & 830.62 & \underline{88.68} & 97.36 & \underline{100.0} & 460.63 \\
    & reasoning\_question       & 93.33 & 98.66 & 99.67 & 836.27 & 77.08 & 94.98 & \underline{99.58} & 545.82 \\
    \cmidrule(lr){2-10}
    & AVG
      & \textbf{88.04}
      & 98.84 & 97.12 & 881.02
      & 76.57 & 95.79
      & \textbf{98.14} & 546.84 \\

    \midrule[0.8pt]
    \multirow{5}{*}{s1.1-7B}
    & analysis\_question        & 74.67 & 99.16 & 94.93 & 573.77 & 63.89 & \underline{99.39} & 91.11 & 299.44 \\
    & instruction\_following    & 47.06 & 98.55 & 81.18 & 2041.02 & 41.38 & \underline{99.19} & 84.83 & 1109.58 \\
    & knowledge\_question       & 80.39 & \underline{100.0} & 94.12 & 848.53 & 84.91 & \underline{100.0} & 96.23 & 329.6 \\
    & reasoning\_question       & 70.0 & 99.28 & 92.67 & 1088.92 & 41.67 & \underline{99.48} & 80.83 & 490.29 \\
    \cmidrule(lr){2-10}
    & AVG       
        & 68.03 & \textbf{99.25}
        & 90.73 & 1138.06 & 57.96 
        & \textbf{99.25}
        & 88.25 & 557.23 \\

    \midrule[0.8pt]
    \multirow{5}{*}{LIMO-7B-reproduced}
    & analysis\_question        & 49.33 & 85.91 & 96.53 & 806.91 & 5.56 & 60.79 & 63.06 & 368.22 \\
    & instruction\_following    & 17.65 & 89.52 & 61.76 & 1633.84 & 41.38 & 90.29 & 71.03 & 1111.96 \\
    & knowledge\_question       & 72.55 & 92.21 & 95.69 & 975.48 & 35.85 & 96.24 & 70.19 & 573.94 \\
    & reasoning\_question       & 56.67 & 81.56 & 94.0 & 1144.3 & 45.83 & 78.0 & 83.33 & 643.32 \\
    \cmidrule(lr){2-10}
    & AVG       & 49.05 & 87.30 & 87.00 & 1140.13 & 32.16 & 81.33 & 71.90 & 674.36 \\
    
    \midrule[0.8pt]
    \multirow{5}{*}{CAC-CoT-7B (Ours)}
    & analysis\_question        & 97.33 & 99.2 & \underline{100.0} & \underline{273.97} & \underline{90.28} & 98.33 & \underline{99.72} & \underline{174.12} \\
    & instruction\_following    & \underline{67.65} & \underline{98.12} & \underline{94.12} & \underline{306.82} & \underline{65.52} & 96.35 & \underline{94.48} & \underline{287.83} \\
    & knowledge\_question       & 84.31 & 99.18 & 96.08 & \underline{256.12} & 84.91 & 99.61 & 97.36 & \underline{177.47} \\
    & reasoning\_question       & \underline{95.00} & \underline{98.67} & \underline{100.0} & \underline{308.13} & \underline{85.42} & 97.49 & \underline{99.58} & \underline{226.16} \\
    \cmidrule(lr){2-10}
    & AVG       
        & 86.07 & 98.79 
        & \textbf{97.55}
        & \textbf{286.26}
        & \textbf{81.53}
        & 97.95 & 97.78 
        & \textbf{216.39}
        \\
    
    \addlinespace[2pt]
      \specialrule{1.5pt}{0pt}{0pt}
    \end{tabular}
    \label{table:s1bench}
}
\end{table*}

\subsection{Experimental Setup}
In this section, we describe our experimental setup and present the main findings of the dual-system benchmark.

\subsubsection{Benchmarks}
For analytical System-2 thinking, we select widely used math-centric datasets of escalating difficulty --- AMC, AIME, GPQA, GSM8K and MATH. Collectively, these benchmarks pose problems that resist easy solutions, driving the model to explore multiple angles and engage in deep reasoning, making them suitable for validation analytical System-2 thinking. For intuitive System-1 thinking, we follow the evaluation metrics of S1-Bench \cite{zhang2025s1bench}, a collection of commonsense and rapid-inference questions that can usually be solved with minimal deliberation. By contrasting performance across these two suites, we can measure whether CAC-CoT preserves fast, concise intuition while enhancing deep analytical skill.

\subsubsection{Baselines}
To contextualize the impact of CAC-CoT, we benchmark against three recent models that rely solely on SFT over Long-CoT data to achieve substantial gains in reasoning performance via System-2 thinking. s1.1 \cite{muennighoff2025s1}, LIMO \cite{ye2025limo}, and Bespoke-Stratos \cite{bespoke_stratos} all rely on SFT over a compact Long-CoT corpus synthesized by powerful reasoning engines (R1, R1-Distill-Qwen-32B, and QwQ, respectively) \citep{guo2025deepseek, deepseekai2025deepseekr1incentivizingreasoningcapability, qwq32b}. Despite the modest data volume, each baseline is reported to extract unexpectedly strong reasoning ability from its target model. s1.1 and Bespoke-Stratos are evaluated exactly as released in huggingface \citep{s11_7bhf, bespoke2024stratoshf}, maintaining full reproducibility. but, LIMO was experimented on the 32B model, and no experimental results were  reported for the 7B model; we therefore reproduce the same method in ours (s1.1 training method), and perform an identical SFT run on Qwen-2.5-7b-Instruct \cite{qwen2.5} checkpoint to ensure a fair comparison.

\subsubsection{Training Details}
\textbf{Training.}  
For fine-tuning, we adopted Qwen-2.5-7B-Instruct as the base model. Our training followed the same hyperparameter configuration used in s1.1 to ensure comparability. Further implementation details, including batch size, optimizer settings, and training duration, are provided in the Appendix \ref{appendix-a}.

\textbf{Datasets.}  
We generated our training data using general-purpose LLMs, Gemini-2.0-Flash and GPT-4o \citep{openai2024gpt4ocard}, cost-effective yet capable frontier models. The performance of our method (Ours) reported in Section 4 is based on synthetic data generated with Gemini-2.0-Flash. To examine whether training signals may have been influenced by the style and bias of a single model, we additionally report results using GPT-4o generated data in Appendix \ref{appendix-c1}. 
During the data generation process, we filtered out instances that exhibited generation errors or failed to follow our specified formatting instructions. As a result, the total usable output was reduced. To ensure sufficient coverage and diversity, we supplemented this data with samples from the LIMO and s1 datasets \citep{ye2025limoreasoning, muennighoff2025s1simpletesttimescaling}. Specifically, we removed any duplicates between LIMO and s1 and excluded all corrupted or invalid generations. After this filtering process, we finalized a training set consisting of \textbf{1,391} examples. Details of the data generation process are provided in Appendix~\ref{appendix-b}.

\subsection{System 1 Thinking}

As summarized in Table \ref{table:s1bench}, our experiments reveal that CAC-CoT consistently achieves strong, state-of-the-art results on System-1 tasks.

We evaluate our method on the S1-Bench suite, which consists of tasks solvable via quick, intuitive reasoning (“System-1” thinking). These tasks generally require little to no step-by-step deduction, so an effective effective should answer accurately without over-elaborating. On this benchmark, it performs on par with or slightly better than the baselines in terms of accuracy: all methods achieve high scores on these easier queries (as expected), but importantly, ours does not sacrifice performance on simple tasks despite its emphasis on CoT. In fact, CAC-CoT attains the highest overall ACC@5 and PASS@1 on S1-Bench (Table~\ref{table:s1bench}), albeit with marginal improvement since the baseline performance is near-saturated.

Crucially, its responses on System-1 tasks are compact and to-the-point, highlighting adaptability. Whereas naive CoT prompting might introduce unnecessary steps or verbiage for trivial questions, our approach generates minimal reasoning --- or sometimes goes straight to the answer --- when extensive explanation is unnecessary. Our average reasoning trace length is the shortest among all evaluated methods on S1-Bench (Table~\ref{table:s1bench}, ART), indicating that it avoids “overthinking” simple problems.

This connector-aware strategy compresses reasoning traces to roughly one-third the length of baselines while preserving --- and in some cases improving --- accuracy. By constraining unnecessary expansion of the CoT, it avoids overthinking pitfalls and shows that instruction-following architectures --- when guided by targeted connector signals --- can overcome the compliance weakness of seen in previous studies on Long CoT reasoning. Empirically, this design delivers over 20\% accuracy gain compared to both s1.1 and LIMO, confirming its strength on straightforward System-1 tasks without sacrificing deliberative performance.

\begin{table*}[t]
\centering
\small
\caption{\textbf{System 2 thinking performance across benchmarks.} Accuracy comparison on five reasoning benchmarks, including AMC23, AIME24, GSM8K, GPQA Diamond and Math500. The row in-between highlights the difference between CAC-CoT-7B (Ours) and the best baseline in each column (orange: negative, blue: positive).}
\resizebox{\textwidth}{!}{%
  \begin{tabular}{lcccccc}
    \toprule
    \textbf{Models}            & \textbf{AMC23} & \textbf{AIME24} & \textbf{GSM8K} & \textbf{GPQA Diamond} & \textbf{Math500} & \textbf{AVG} \\
    \midrule
    Qwen2.5-7B-Instruct        & 55.00          & 6.67       &79.98     & 33.84                 & 75.00             & 50.09        \\
    s1.1-7B                    & 55.00          & 13.33      &90.67     & 39.39                 & 79.40             & 55.55        \\[1.0pt]
    LIMO-7B-reproduced         & 57.50          & 13.33       &88.55    & 35.35                 & 78.20            & 54.58        \\[1.0pt]
    Bespoke-Stratos-7B         & 52.50          & 23.33      &88.25     & 43.94                 & 80.20             & 57.64        \\
    \midrule
                               & \textcolor{orange}{\scriptsize–5.0} & \textcolor{blue}{\scriptsize+3.33} & \textcolor{blue}{\scriptsize+5.39} & \textcolor{blue}{\scriptsize+4.54} & \textcolor{orange}{\scriptsize–7.0} & \textcolor{blue}{\scriptsize+0.26} \\[-0.3ex]
    CAC-CoT-7B (Ours)          & 50.00          & 10.00        &85.37   & 38.38                 & 68.00            & 50.35        \\[1.2pt]
    \bottomrule
  \end{tabular}%
}
\label{tab:performance_comparison}
\end{table*}

\begin{figure}[t]
   \includegraphics[width=1\columnwidth]{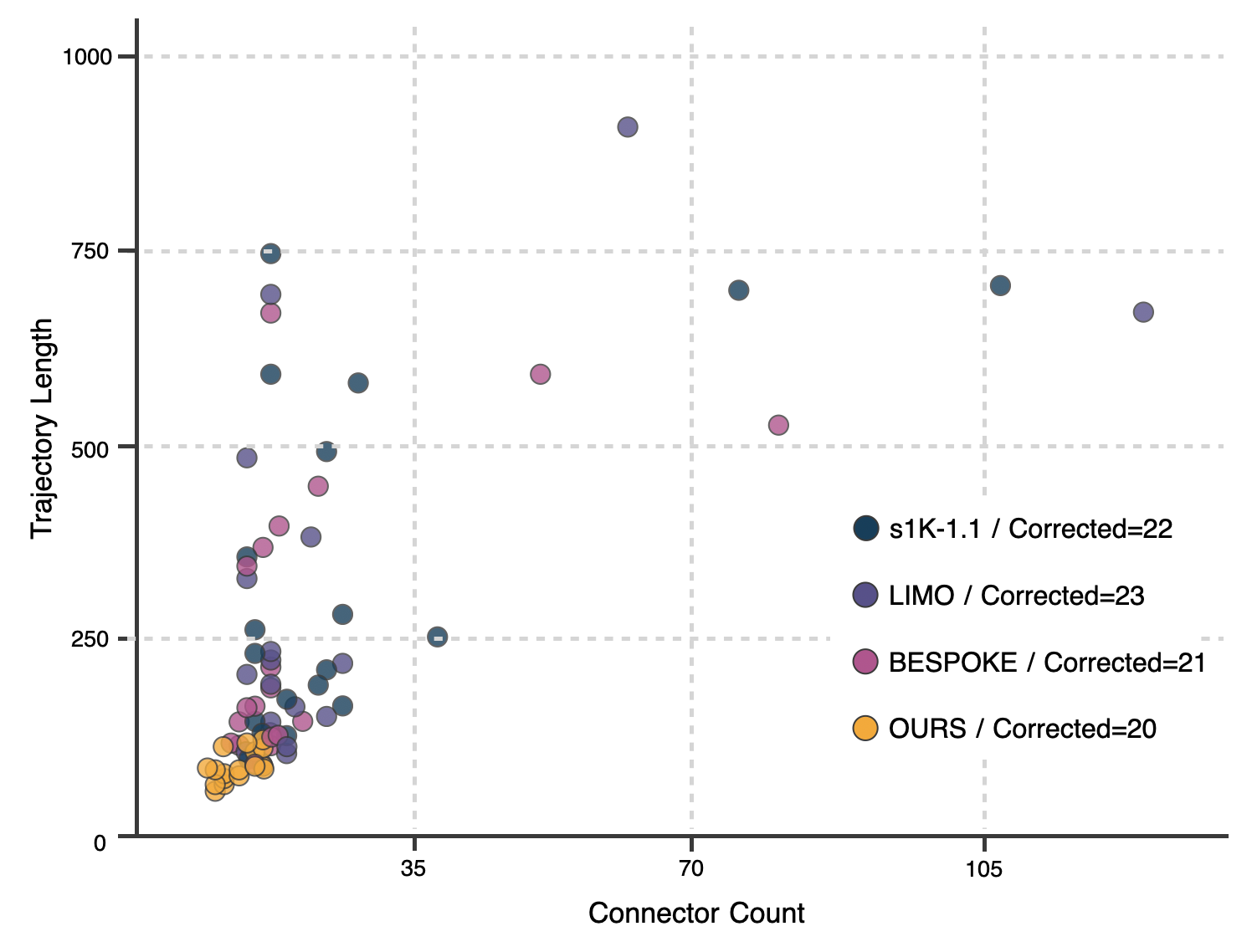}
  \caption{\textbf{Compactness of Chain-of-Thought Traces by Model.}
Scatter plot of reasoning-trace length (y-axis) versus connector count (x-axis) for AMC23 outputs. Points farther toward the lower-left denote shorter, more compact traces.}
  \label{fig:reasoningtraceanalysis}
\end{figure}

\subsection{System 2 Thinking}

CAC-CoT maintains strong performance on System-2 benchmarks, demonstrating comparable reasoning ability to prior Long-CoT baselines. While there is a slight performance penalty relative to baseline models --- whose results are detailed in Table~\ref{tab:performance_comparison} --- the overall degradation is minor, and Ours successfully preserves analytical reasoning capabilities without relying on large-scale reasoning-specialized traces.

A key distinction lies in the construction of our training data. Unlike existing approaches that depend heavily on dedicated reasoning models to generate Long-CoT corpora --- often resulting in verbose outputs with excessive connector usage --- it employs a connector-aware generation strategy based on Gemini-2.0-flash, a general-purpose but instruction-aligned model. This allows us to retain the reasoning efficacy of connector-based CoT without inflating either the length of reasoning traces or the frequency of connector insertions. As shown in Figure~\ref{fig:reasoningtraceanalysis}, our outputs are significantly more compact, with the CAC-CoT cluster consistently positioned in the lower-left region, indicating fewer connectors and greater efficiency than competing models. In summary, CAC-CoT offers a principled trade-off: it slightly underperforms the strongest reasoning-optimized baselines, yet does so with far more economical prompting and without the bloat of overly long reasoning chains. This makes it a compelling choice for efficient, scalable reasoning under constrained data and compute regimes.

\begin{table}[!hbt]
  \centering
  \caption{\textbf{Connector usage comparison between open-source LongCoT corpora and our corpus.}
Our corpus yields significantly shorter reasoning traces with lower connector density compared to widely used open-source LongCoT datasets.
\textbf{Len}: average trajectory length (tokens).
\textbf{Conn/1K}: connectors per 1{,}000 tokens.
\textbf{\# Samples}: number of samples.}
  \small
  \resizebox{\columnwidth}{!}{%
    \begin{tabular}{l|ccc} % ← 4열에 맞게
      \toprule
      \textbf{Dataset} & \textbf{Len} & \textbf{Conn/1K} & \textbf{\# Samples} \\
      \midrule
      \textsc{s1k-1.1} & 9291.62 & 5.55 & 1k \\
      \textsc{limo}    & 6984.09 & 2.97 & 0.8k \\
      \textsc{bespoke} & 4452.22 & 5.13 & 16.7k \\
      \textsc{ours}    & 1843.43 & 2.65 & 1.4k \\
      \bottomrule
    \end{tabular}%
  }
  \label{table:connector_ultra}
\end{table}

\subsection{Connector Usage}

% \begin{figure}[t]
%     \includegraphics[width=1\columnwidth]{figures/correct_incorrect.png}
%   \caption{\textbf{Impact of connector annotation strategy on System-1 and System-2 thinking performance.}
% \textcolor{red}{Top}: System-1 metrics for models trained with connectors on all, only incorrect, or only correct reasoning. Right y-axis shows ART (average reasoning trajectory length in tokens).
% \textcolor{blue}{Bottom}: System-2 accuracy under the same settings.}
%   \label{fig:datacorrectness}
% \end{figure}
To better understand the efficiency and structure of reasoning under CAC-CoT, we jointly analyze the distribution of connector usage and output length across both the training data and inference outputs, comparing against baselines (Table~\ref{table:connector_ultra} , Figure~\ref{fig:reasoningtraceanalysis}).
These two aspects --- connector count and sequence length --- serve as proxies for reasoning verbosity and structural control, and offer insight into how different models manage the trade-off between expressiveness and conciseness.
Table~\ref{table:connector_ultra} summarizes connector usage and token lengths for our training corpora versus the baselines. The Bespoke dataset averages approximately 4500 tokens per reasoning trace with 5 connectors per 1000 tokens, while s1K averages around 9000 tokens and 5.5 connectors per 1000 tokens. In contrast, our connector-aware data exhibits a far more controlled structure, with an average trace length of only 1850 tokens and just 2.5 connectors per 1000 tokens. This demonstrates that our data construction yields an efficient reasoning corpus, which in turn benefits inference. Figure~\ref{fig:reasoningtraceanalysis} shows connector counts and token lengths at inference time. As evident, the compact structure of the training data carries over to outputs: on average the model uses only about 20 connectors and produces fewer than 200 tokens per response. Baseline models, by comparison, sometimes exceed 500 tokens or include over 70 connectors. Together with the accuracy efficiency improvements reported in Sections 4.2 and 4.3, these results confirm that our connector-aware approach delivers both concise and efficient reasoning.

Additionally, we perform a connector-aware qualitative analysis on the generated training data. Table~\ref{table:usageconnector} displays how the data patterns are structured before and after each connector. The first is the reflection connector, which reflects known mechanisms such as self-reflection and backtracking. Theses connectors encourage exploratory or verification-oriented reasoning. The second type is the confidence connector, which signals confirmation of the current reasoning path or indicates termination of the reasoning process. The text segments before and after each connector reflect its intended function, demonstrating how the logic is shaped in alignment with the connector type.

The connector generation procedure is described in Appendix \ref{appendix-c2}, and additional ablation studies on connectors are provided therein.

\subsection{Impact of Compactness and Connector}

To separate the contributions of compactness and connector usage, we conducted an additional study with two separate synthetic datasets: one generated using only connector-aware (Only Connector-Aware) and the other using only compactness (Only Compact). The results are shown in Table~\ref{tab:restrictions}

When excluding connectors and applying only compactness, the overall length of synthetic data was greatly reduced. However, the System-2 average score dropped by about 10\% compared to the base score. This indicates that the self-reflection and confidence signals provided by connector-aware reasoning substantially enhance reasoning performance, particularly in challenging System-2 tasks. Interestingly, applying compactness alone improved System-1 performance, suggesting that compactness can be effective for simpler tasks, whereas connectors provide essential scaffolding for more complex reasoning scenarios.

\clearpage
\begin{table*}[t]
  \centering
  \caption{\textbf{Usage of Reflection and Confidence Connectors.} The \textcolor{cyan}{light-blue} highlighted phrases represent reflection connectors, which typically lead to revalidation or exploration of alternative reasoning paths. In contrast, the \textcolor{lightorange}{light-orange} highlighted phrases indicate confidence connectors, which are usually followed by concretization of reasoning, confirmation of the current logic, or a transition toward final conclusions.}
  \begin{tabularx}{\textwidth}{>{\bfseries}l >{\raggedright\arraybackslash}X}

    \toprule
    Reflection Connector & \textcolor{cyan}{I might have overlooked something. Let’s pause and rethink this.} The pattern doesn't seem obvious. Let's re-evaluate that approach. Instead of trying to find a pattern, let's try to show that there are infinitely many losing positions. \\
    \\
    & \textcolor{cyan}{This doesn’t lead where we want it to. Let's pause and rethink this.} Consider the Sprague-Grundy theorem. Let $g(n)$ be the Grundy value of $n$ pebbles.\\
    \\
    & \textcolor{cyan}{However, this might not be the right path because} the problem says that $\Delta E_{rms}$ depends on the rms value of the component of velocity along the line of sight. We used the full kinetic energy expression, but we should have used only one component. \textcolor{cyan}{Let's re-evaluate that approach.}\\
    \\
    & Now let's move to the equation for $y(t)$.
$y(t) = \int_0^t e^{-2(t-s)} \{2x(s) + 3y(s)\} ds$.
$y(t) = e^{-2t} * (2x(t) + 3y(t))$.
Taking the Laplace transform:
$Y(p) = \mathcal{L}\{e^{-2t}\} \mathcal{L}\{2x(t) + 3y(t)\}$.
$Y(p) = \frac{1}{p+2} (2X(p) + 3Y(p))$.
$Y(p) = \frac{1}{p+2} (2(\frac{2}{p} - \frac{1}{p+1}) + 3Y(p))$.
$Y(p) (1 - \frac{3}{p+2}) = \frac{2}{p+2} (\frac{2}{p} - \frac{1}{p+1})$.
$Y(p) (\frac{p+2-3}{p+2}) = \frac{2}{p+2} (\frac{2(p+1)-p}{p(p+1)})$.
$Y(p) (\frac{p-1}{p+2}) = \frac{2}{p+2} (\frac{p+2}{p(p+1)})$.
$Y(p) (\frac{p-1}{p+2}) = \frac{2}{p+2} \frac{p+2}{p(p+1)}$.
$Y(p) = \frac{2}{p(p+1)} \frac{p+2}{p+2} \frac{1}{\frac{p-1}{p+2}}$.
$Y(p) = \frac{2}{p(p+1)} \frac{p+2}{p-1}$.

\textcolor{cyan}{Hmm, that might be a dead end.}
$Y(p) = \frac{2(p+2)}{p(p+1)(p-1)}$.
Using partial fractions:
$\frac{2(p+2)}{p(p+1)(p-1)} = \frac{A}{p} + \frac{B}{p+1} + \frac{C}{p-1}$.
$2(p+2) = A(p+1)(p-1) + Bp(p-1) + Cp(p+1)$.
$2p+4 = A(p^2-1) + B(p^2-p) + C(p^2+p)$.
$2p+4 = (A+B+C)p^2 + (-B+C)p - A$.
    \\

    \midrule
    
    Confidence Connector   & \textcolor{lightorange}{Now that’s convincing, it really does.} Let us assume that for some $m$, $4m+2$ is a losing position. Now we want to show that we can find infinitely many numbers of the form $4m+2$ are losing. \\
    \\
    & \textcolor{lightorange}{Wow, that actually makes a lot of sense now.} Let $a_n$ be a losing position if and only if $g(a_n) = 0$. We know $g(0) = 0$ and $g(2) = 0$, so $0$ and $2$ are losing positions. We want to find infinitely many losing positions. Let's find $g(n)$ for $n \le 20$. \\
    \\
    & \textcolor{lightorange}{Wow, that actually makes a lot of sense now.}

We have $N = 2^{2011} \cdot 2025078 \pmod{1000}$.
We have $2025078 \equiv 78 \pmod{1000}$.
Also, we have $2^{2011} \equiv 2^{11} \pmod{1000}$, since $2^{100} \equiv 376 \pmod{1000}$.
Also $2^{10} = 1024 \equiv 24 \pmod{1000}$.
Then $2^{11} = 2048 \equiv 48 \pmod{1000}$.
So, $N \equiv 48 \cdot 78 \equiv 3744 \equiv 744 \pmod{1000}$. \textcolor{lightorange}{Everything fits together nicely.}\\
    \\
    & \textcolor{lightorange}{Looks consistent with earlier steps.} I think this is the correct approach using only one degree of freedom to determine the temperature. \\

    \bottomrule
  \end{tabularx}
  \label{table:usageconnector}
\end{table*}
\clearpage

% 본문 중 원하는 위치
\begin{strip}
  \centering
  \small
  \captionof{table}{\textbf{Comparison across restricted variants.}
  Comparison of synthetic data lengths under individual constraints on connector-awareness and compactness, with corresponding impact on System-2 (S2) and System-1 (S1) performance (AVG, ACC@5, ART).}
  \label{tab:restrictions}
  \resizebox{\textwidth}{!}{%
    \begin{tabular}{lcccccc}
      \toprule
      \textbf{Methods} & \textbf{Avg Token Len.} & \textbf{S2 AVG} & \textbf{S2 ART} & \textbf{S2 ART (Incorrect)} & \textbf{S1 AVG} & \textbf{S1 ART} \\
      \midrule
      Only Connector-Aware & 1,856 & 50.166 & 1973.2 & 2585 & 85.07 & 247 \\
      Only Compactness     & 1,228 & 45.42  & 1053.2 & 1242 & 92.65 & 104 \\
      Connector-Aware + Compactness (CAC-CoT) & 1,843 & 50.35 & 1681.8 & 2125 & 83.92 & 252 \\
      \bottomrule
    \end{tabular}%
  }
\end{strip}

Comparing Ours (Connector-Aware + Compact) with Only Connector-Aware revealed almost no difference in average System-2 scores, implying that Only Connector-Aware is already sufficient to maintain reasoning quality. Nevertheless, ART decreased by approximately 300 tokens in System-2 tasks and by about 450 tokens in incorrect cases. This demonstrates that compactness reduces unnecessary tokens and improves reasoning efficiency, particularly by preventing overly long and unsuccessful reasoning paths.

\section{Conclusion}

We present a prompt-based, connector-aware compact chain-of-thought framework that automatically generates high-quality reasoning traces with general-purpose LLMs such as Gemini-2.0-Flash and GPT-4o and trains models exclusively through SFT. By enforcing depth limits and explicit connector cues, our method produces concise yet coherent rationales that improve both reliability and efficiency. Extensive experiments confirm its effectiveness: the approach consistently surpasses strong baselines across intuitive System-1 and analytical System-2 benchmarks while operating with significantly lower computational overhead.
% \section{Bib\TeX{} Files}
% \label{sec:bibtex}

% Unicode cannot be used in Bib\TeX{} entries, and some ways of typing special characters can disrupt Bib\TeX's alphabetization. The recommended way of typing special characters is shown in Table~\ref{tab:accents}.

% Please ensure that Bib\TeX{} records contain DOIs or URLs when possible, and for all the ACL materials that you reference.
% Use the \verb|doi| field for DOIs and the \verb|url| field for URLs.
% If a Bib\TeX{} entry has a URL or DOI field, the paper title in the references section will appear as a hyperlink to the paper, using the hyperref \LaTeX{} package.

\section*{Limitations}

The findings reported here rest on notable constraints. First, all experiments were performed with a single model family, Qwen, leaving open the question of whether comparable improvements would materialize for other widely used backbones such as LLaMA or Gemma.

Second, no quantitative comparison was made against RL-based models. While our study demonstrates a methodology that efficiently and effectively enhances reasoning in general-purpose Language Models such as Qwen2.5-7B-Instruct, recent advances show that RL-based models are achieving increasingly strong reasoning performance. Whether our approach can be optimized for RL-based models remains an open question. Given the differing characteristics between general-purpose and RL-based Language Models, adapting the proposed method for RL-based settings represents an important direction for future work.

Third, while categorizing connectors into confidence and reflection types may be considered a strength of our methodology, the construction of the connector pool itself was based on heuristic rules. This design choice, though practical, introduces a potential source of bias and limits the systematic generalizability of the results. Future work could explore more principled or data-driven approaches for defining connector categories to enhance the robustness of evaluation.

Broadening the range of backbone models and extending evaluation to RL-based models would therefore provide a more rigorous test of the generalizability, robustness, and transferability of the proposed approach.

% \section*{Acknowledgments}

% This document has been adapted
% by Steven Bethard, Ryan Cotterell and Rui Yan
% from the instructions for earlier ACL and NAACL proceedings, including those for
% ACL 2019 by Douwe Kiela and Ivan Vuli\'{c},
% NAACL 2019 by Stephanie Lukin and Alla Roskovskaya,
% ACL 2018 by Shay Cohen, Kevin Gimpel, and Wei Lu,
% NAACL 2018 by Margaret Mitchell and Stephanie Lukin,
% Bib\TeX{} suggestions for (NA)ACL 2017/2018 from Jason Eisner,
% ACL 2017 by Dan Gildea and Min-Yen Kan,
% NAACL 2017 by Margaret Mitchell,
% ACL 2012 by Maggie Li and Michael White,
% ACL 2010 by Jing-Shin Chang and Philipp Koehn,
% ACL 2008 by Johanna D. Moore, Simone Teufel, James Allan, and Sadaoki Furui,
% ACL 2005 by Hwee Tou Ng and Kemal Oflazer,
% ACL 2002 by Eugene Charniak and Dekang Lin,
% and earlier ACL and EACL formats written by several people, including
% John Chen, Henry S. Thompson and Donald Walker.
% Additional elements were taken from the formatting instructions of the \emph{International Joint Conference on Artificial Intelligence} and the \emph{Conference on Computer Vision and Pattern Recognition}.


\begin{thebibliography}{36}
\providecommand{\natexlab}[1]{#1}

\bibitem[{Adarsh et~al.(2024)Adarsh, Shridhar, Gulcehre, Monath, and Sachan}]{adarsh2024siked}
Shivam Adarsh, Kumar Shridhar, Caglar Gulcehre, Nicholas Monath, and Mrinmaya Sachan. 2024.
\newblock Siked: Self-guided iterative knowledge distillation for mathematical reasoning.
\newblock \emph{arXiv preprint arXiv:2410.18574}.

\bibitem[{Anthropic(2025)}]{anthropic2025claude37}
Anthropic. 2025.
\newblock Claude 3.7 sonnet system card.
\newblock \url{https://assets.anthropic.com/m/785e231869ea8b3b/original/claude-3-7-sonnet-system-card.pdf}.

\bibitem[{Chen et~al.(2025)Chen, Qin, Liu, Peng, Guan, Wang, Hu, Zhou, Gao, and Che}]{chen2025towards}
Qiguang Chen, Libo Qin, Jinhao Liu, Dengyun Peng, Jiannan Guan, Peng Wang, Mengkang Hu, Yuhang Zhou, Te~Gao, and Wanxiang Che. 2025.
\newblock Towards reasoning era: A survey of long chain-of-thought for reasoning large language models.
\newblock \emph{arXiv preprint arXiv:2503.09567}.

\bibitem[{Chen et~al.(2024)Chen, Xu, Liang, He, Pang, Yu, Song, Liu, Zhou, Zhang et~al.}]{chen2024not}
Xingyu Chen, Jiahao Xu, Tian Liang, Zhiwei He, Jianhui Pang, Dian Yu, Linfeng Song, Qiuzhi Liu, Mengfei Zhou, Zhuosheng Zhang, and 1 others. 2024.
\newblock Do not think that much for 2+ 3=? on the overthinking of o1-like llms.
\newblock \emph{arXiv preprint arXiv:2412.21187}.

\bibitem[{DeepSeek-AI(2025)}]{deepseekai2025deepseekr1incentivizingreasoningcapability}
DeepSeek-AI. 2025.
\newblock \href {https://arxiv.org/abs/2501.12948} {Deepseek-r1: Incentivizing reasoning capability in llms via reinforcement learning}.
\newblock \emph{Preprint}, arXiv:2501.12948.

\bibitem[{Google(2025)}]{google2025gemini25}
Google. 2025.
\newblock Gemini 2.5 pro preview model card.
\newblock \url{https://storage.googleapis.com/model-cards/documents/gemini-2.5-pro-preview.pdf}.
\newblock Also published on the Google AI and DeepMind Blogs.

\bibitem[{Guo et~al.(2025)Guo, Yang, Zhang, Song, Zhang, Xu, Zhu, Ma, Wang, Bi et~al.}]{guo2025deepseek}
Daya Guo, Dejian Yang, Haowei Zhang, Junxiao Song, Ruoyu Zhang, Runxin Xu, Qihao Zhu, Shirong Ma, Peiyi Wang, Xiao Bi, and 1 others. 2025.
\newblock Deepseek-r1: Incentivizing reasoning capability in llms via reinforcement learning.
\newblock \emph{arXiv preprint arXiv:2501.12948}.

\bibitem[{Jaech et~al.(2024)Jaech, Kalai, Lerer, Richardson, El-Kishky, Low, Helyar, Madry, Beutel, Carney et~al.}]{jaech2024openai}
Aaron Jaech, Adam Kalai, Adam Lerer, Adam Richardson, Ahmed El-Kishky, Aiden Low, Alec Helyar, Aleksander Madry, Alex Beutel, Alex Carney, and 1 others. 2024.
\newblock Openai o1 system card.
\newblock \emph{arXiv preprint arXiv:2412.16720}.

\bibitem[{Kahneman(2011)}]{kahneman2011thinking}
Daniel Kahneman. 2011.
\newblock \emph{Thinking, fast and slow}.
\newblock macmillan.

\bibitem[{Kannengiesser and Gero(2019)}]{kannengiesser2019design}
Udo Kannengiesser and John~S Gero. 2019.
\newblock Design thinking, fast and slow: A framework for kahneman’s dual-system theory in design.
\newblock \emph{Design Science}, 5:e10.

\bibitem[{Labs(2024)}]{bespoke2024stratoshf}
Bespoke Labs. 2024.
\newblock \href {https://huggingface.co/bespokelabs/Bespoke-Stratos-7B} {Bespoke-stratos-7b}.
\newblock Accessed: 2025-05-20.

\bibitem[{Labs(2025)}]{bespoke_stratos}
Bespoke Labs. 2025.
\newblock Bespoke-stratos: The unreasonable effectiveness of reasoning distillation.
\newblock https://www.bespokelabs.ai/blog/bespoke-stratos-the-unreasonable-effectiveness-of-reasoning-distillation.
\newblock Accessed: 2025-01-22.

\bibitem[{Li et~al.(2025{\natexlab{a}})Li, Cao, Griggs, Liu, Mo, Tang, Hegde, Hakhamaneshi, Patil, Zaharia et~al.}]{li2025llm}
Dacheng Li, Shiyi Cao, Tyler Griggs, Shu Liu, Xiangxi Mo, Eric Tang, Sumanth Hegde, Kourosh Hakhamaneshi, Shishir~G Patil, Matei Zaharia, and 1 others. 2025{\natexlab{a}}.
\newblock Llms can easily learn to reason from demonstrations structure, not content, is what matters!
\newblock \emph{arXiv preprint arXiv:2502.07374}.

\bibitem[{Li et~al.(2025{\natexlab{b}})Li, Zou, and Liu}]{li2025limr}
Xuefeng Li, Haoyang Zou, and Pengfei Liu. 2025{\natexlab{b}}.
\newblock Limr: Less is more for rl scaling.
\newblock \emph{arXiv preprint arXiv:2502.11886}.

\bibitem[{Li et~al.(2025{\natexlab{c}})Li, Yue, Xu, Jiang, Niu, Lin, Ramasubramanian, and Poovendran}]{li2025small}
Yuetai Li, Xiang Yue, Zhangchen Xu, Fengqing Jiang, Luyao Niu, Bill~Yuchen Lin, Bhaskar Ramasubramanian, and Radha Poovendran. 2025{\natexlab{c}}.
\newblock Small models struggle to learn from strong reasoners.
\newblock \emph{arXiv preprint arXiv:2502.12143}.

\bibitem[{Madaan et~al.(2023)Madaan, Tandon, Gupta, Hallinan, Gao, Wiegreffe, Alon, Dziri, Prabhumoye, Yang et~al.}]{madaan2023self}
Aman Madaan, Niket Tandon, Prakhar Gupta, Skyler Hallinan, Luyu Gao, Sarah Wiegreffe, Uri Alon, Nouha Dziri, Shrimai Prabhumoye, Yiming Yang, and 1 others. 2023.
\newblock Self-refine: Iterative refinement with self-feedback.
\newblock \emph{Advances in Neural Information Processing Systems}, 36:46534--46594.

\bibitem[{Muennighoff et~al.(2025{\natexlab{a}})Muennighoff, Yang, Shi, Li, Fei-Fei, Hajishirzi, Zettlemoyer, Liang, Cand{\`e}s, and Hashimoto}]{muennighoff2025s1}
Niklas Muennighoff, Zitong Yang, Weijia Shi, Xiang~Lisa Li, Li~Fei-Fei, Hannaneh Hajishirzi, Luke Zettlemoyer, Percy Liang, Emmanuel Cand{\`e}s, and Tatsunori Hashimoto. 2025{\natexlab{a}}.
\newblock s1: Simple test-time scaling.
\newblock \emph{arXiv preprint arXiv:2501.19393}.

\bibitem[{Muennighoff et~al.(2025{\natexlab{b}})Muennighoff, Yang, Shi, Li, Fei-Fei, Hajishirzi, Zettlemoyer, Liang, Candès, and Hashimoto}]{muennighoff2025s1simpletesttimescaling}
Niklas Muennighoff, Zitong Yang, Weijia Shi, Xiang~Lisa Li, Li~Fei-Fei, Hannaneh Hajishirzi, Luke Zettlemoyer, Percy Liang, Emmanuel Candès, and Tatsunori Hashimoto. 2025{\natexlab{b}}.
\newblock \href {https://arxiv.org/abs/2501.19393} {s1: Simple test-time scaling}.
\newblock \emph{Preprint}, arXiv:2501.19393.

\bibitem[{Pang et~al.(2024)Pang, Yuan, He, Cho, Sukhbaatar, and Weston}]{pang2024iterative}
Richard~Yuanzhe Pang, Weizhe Yuan, He~He, Kyunghyun Cho, Sainbayar Sukhbaatar, and Jason Weston. 2024.
\newblock Iterative reasoning preference optimization.
\newblock \emph{Advances in Neural Information Processing Systems}, 37:116617--116637.

\bibitem[{Rein et~al.(2024)Rein, Hou, Stickland, Petty, Pang, Dirani, Michael, and Bowman}]{rein2024gpqa}
David Rein, Betty~Li Hou, Asa~Cooper Stickland, Jackson Petty, Richard~Yuanzhe Pang, Julien Dirani, Julian Michael, and Samuel~R Bowman. 2024.
\newblock Gpqa: A graduate-level google-proof q\&a benchmark.
\newblock In \emph{First Conference on Language Modeling}.

\bibitem[{Shen et~al.(2025)Shen, Huang, Zhao, Liu, Zheng, and Zhu}]{shen2025long}
Si~Shen, Fei Huang, Zhixiao Zhao, Chang Liu, Tiansheng Zheng, and Danhao Zhu. 2025.
\newblock Long is more important than difficult for training reasoning models.
\newblock \emph{arXiv preprint arXiv:2503.18069}.

\bibitem[{Snell et~al.(2024)Snell, Lee, Xu, and Kumar}]{snell2024scalingllmtesttimecompute}
Charlie Snell, Jaehoon Lee, Kelvin Xu, and Aviral Kumar. 2024. 
\newblock Scaling LLM Test-Time Compute Optimally can be More Effective than Scaling Model Parameters. 
\newblock \emph{arXiv preprint arXiv:2408.03314}.


\bibitem[{Shinn et~al.(2023)Shinn, Cassano, Gopinath, Narasimhan, and Yao}]{shinn2023reflexion}
Noah Shinn, Federico Cassano, Ashwin Gopinath, Karthik Narasimhan, and Shunyu Yao. 2023.
\newblock Reflexion: Language agents with verbal reinforcement learning.
\newblock \emph{Advances in Neural Information Processing Systems}, 36:8634--8652.

\bibitem[{{SimpleScaling Team}(2024)}]{s11_7bhf}
{SimpleScaling Team}. 2024.
\newblock s1.1-7b model on hugging face.
\newblock \url{https://huggingface.co/simplescaling/s1.1-7B}.
\newblock Accessed: 2025-05-19.

\bibitem[{Sui et~al.(2025)Sui, Chuang, Wang, Zhang, Zhang, Yuan, Liu, Wen, Zhong, Chen et~al.}]{sui2025stop}
Yang Sui, Yu-Neng Chuang, Guanchu Wang, Jiamu Zhang, Tianyi Zhang, Jiayi Yuan, Hongyi Liu, Andrew Wen, Shaochen Zhong, Hanjie Chen, and 1 others. 2025.
\newblock Stop overthinking: A survey on efficient reasoning for large language models.
\newblock \emph{arXiv preprint arXiv:2503.16419}.

\bibitem[{Swayamdipta et~al.(2020)Swayamdipta, Schwartz, Lourie, Wang, Hajishirzi, Smith, and Choi}]{swayamdipta2020dataset}
Swabha Swayamdipta, Roy Schwartz, Nicholas Lourie, Yizhong Wang, Hannaneh Hajishirzi, Noah~A Smith, and Yejin Choi. 2020.
\newblock Dataset cartography: Mapping and diagnosing datasets with training dynamics.
\newblock \emph{arXiv preprint arXiv:2009.10795}.

\bibitem[{Team(2024)}]{qwen2.5}
Qwen Team. 2024.
\newblock \href {https://qwenlm.github.io/blog/qwen2.5/} {Qwen2.5: A party of foundation models}.

\bibitem[{Team(2025)}]{qwq32b}
Qwen Team. 2025.
\newblock \href {https://qwenlm.github.io/blog/qwq-32b/} {Qwq-32b: Embracing the power of reinforcement learning}.

\bibitem[{Wei et~al.(2022)Wei, Wang, Schuurmans, Bosma, Xia, Chi, Le, Zhou et~al.}]{wei2022chain}
Jason Wei, Xuezhi Wang, Dale Schuurmans, Maarten Bosma, Fei Xia, Ed~Chi, Quoc~V Le, Denny Zhou, and 1 others. 2022.
\newblock Chain-of-thought prompting elicits reasoning in large language models.
\newblock \emph{Advances in neural information processing systems}, 35:24824--24837.

\bibitem[{Wu et~al.(2025)Wu, Wang, Du, Jegelka, and Wang}]{wu2025more}
Yuyang Wu, Yifei Wang, Tianqi Du, Stefanie Jegelka, and Yisen Wang. 2025.
\newblock When more is less: Understanding chain-of-thought length in llms.
\newblock \emph{arXiv preprint arXiv:2502.07266}.

\bibitem[{Xu et~al.(2025{\natexlab{a}})Xu, Peng, Awadalla, Chen, Chen, Gao, Kim, Li, Ren, Shen et~al.}]{xu2025phi}
Haoran Xu, Baolin Peng, Hany Awadalla, Dongdong Chen, Yen-Chun Chen, Mei Gao, Young~Jin Kim, Yunsheng Li, Liliang Ren, Yelong Shen, and 1 others. 2025{\natexlab{a}}.
\newblock Phi-4-mini-reasoning: Exploring the limits of small reasoning language models in math.
\newblock \emph{arXiv preprint arXiv:2504.21233}.

\bibitem[{Xu et~al.(2025{\natexlab{b}})Xu, Wu, Wang, Li, Zheng, Chen, Hu, Kang, Ji, Zhang et~al.}]{xu2025redstar}
Haotian Xu, Xing Wu, Weinong Wang, Zhongzhi Li, Da~Zheng, Boyuan Chen, Yi~Hu, Shijia Kang, Jiaming Ji, Yingying Zhang, and 1 others. 2025{\natexlab{b}}.
\newblock Redstar: Does scaling long-cot data unlock better slow-reasoning systems?
\newblock \emph{arXiv preprint arXiv:2501.11284}.

\bibitem[{Yao et~al.(2023{\natexlab{a}})Yao, Yu, Zhao, Shafran, Griffiths, Cao, and Narasimhan}]{yao2023tree}
Shunyu Yao, Dian Yu, Jeffrey Zhao, Izhak Shafran, Tom Griffiths, Yuan Cao, and Karthik Narasimhan. 2023{\natexlab{a}}.
\newblock Tree of thoughts: Deliberate problem solving with large language models.
\newblock \emph{Advances in neural information processing systems}, 36:11809--11822.

\bibitem[{Yao et~al.(2023{\natexlab{b}})Yao, Li, and Zhao}]{yao2023beyond}
Yao Yao, Zuchao Li, and Hai Zhao. 2023{\natexlab{b}}.
\newblock Beyond chain-of-thought, effective graph-of-thought reasoning in language models.
\newblock \emph{arXiv preprint arXiv:2305.16582}.

\bibitem[{Ye et~al.(2025{\natexlab{a}})Ye, Huang, Xiao, Chern, Xia, and Liu}]{ye2025limo}
Yixin Ye, Zhen Huang, Yang Xiao, Ethan Chern, Shijie Xia, and Pengfei Liu. 2025{\natexlab{a}}.
\newblock Limo: Less is more for reasoning.
\newblock \emph{arXiv preprint arXiv:2502.03387}.

\bibitem[{Ye et~al.(2025{\natexlab{b}})Ye, Huang, Xiao, Chern, Xia, and Liu}]{ye2025limoreasoning}
Yixin Ye, Zhen Huang, Yang Xiao, Ethan Chern, Shijie Xia, and Pengfei Liu. 2025{\natexlab{b}}.
\newblock \href {https://arxiv.org/abs/2502.03387} {Limo: Less is more for reasoning}.
\newblock \emph{Preprint}, arXiv:2502.03387.

\bibitem[{Zhang et~al.(2025)Zhang, Nie, Zhang, Zhang, and Liu}]{zhang2025s1bench}
Wenyuan Zhang, Shuaiyi Nie, Xinghua Zhang, Zefeng Zhang, and Tingwen Liu. 2025.
\newblock S1-bench: A simple benchmark for evaluating system 1 thinking capability of large reasoning models.
\newblock \emph{arXiv preprint arXiv:2504.10368}.

\bibitem[{OpenAI et~al.(2024)}]{openai2024gpt4ocard}
OpenAI, Aaron Hurst, Adam Lerer, Adam P. Goucher, Adam Perelman, Aditya Ramesh, 
\textit{et~al.} 2024.  
\newblock GPT-4o System Card.  
\newblock \emph{arXiv preprint arXiv:2410.21276}.  

\bibitem[{Schiffrin(1987)}]{schiffrin1987discourse}
Schiffrin, Deborah. 1987.  
\newblock \emph{Discourse Markers}.  
\newblock Cambridge University Press. 


\end{thebibliography}

\appendix

\label{sec:appendix}

\clearpage

\section{Implementation and Training Configuration}
\label{appendix-a}
To enable a fair comparison with existing baselines, we adopt the same training setup used for s1.1-7B \citep{s11_7bhf, muennighoff2025s1} and fine-tune our model on the proposed reasoning dataset. Specifically, we use Qwen2.5-7B-Instruct with the same hyperparameters as in prior work. The model is trained for 5 epochs with a batch size of 16 and a gradient accumulation step of 4, using four NVIDIA A100 80GB GPUs. However, unlike previous datasets, our reasoning data contains relatively long samples (approximately 2,000 tokens on average), and thus we set the block size to 4,000 to ensure full context coverage during training. We also reproduce the LIMO baseline (denoted as LIMO-Reproduce) under the same hyperparameter settings for performance comparison. The only difference is the block size, which we set to 13,000 due to the length of the input and GPU memory constraints. The full training configuration is summarized as follows:

\begin{table}[h]
\centering
\caption{\textbf{Training Configuration}}
\resizebox{1\columnwidth}{!}{
\begin{tabular}{ll}
\toprule
\textbf{Parameter} & \textbf{Value} \\
\midrule
Optimizer & AdamW ($\beta_1 = 0.9$, $\beta_2 = 0.95$) \\
Scheduler & Cosine \\
Learning Rate & $1 \times 10^{-5}$ \\
Per Device Train Batch Size & 1 \\
Gradient Accumulation Steps & 4 \\
Block Size & 4000 \\
Weight Decay & $1 \times 10^{-4}$ \\
Hardware & 4$\times$NVIDIA A100 (80GB) \\
\bottomrule
\end{tabular}
}
\end{table}

\section{Training Datasets}
\label{appendix-b}
\begin{figure}[t]
   \includegraphics[width=1\columnwidth]{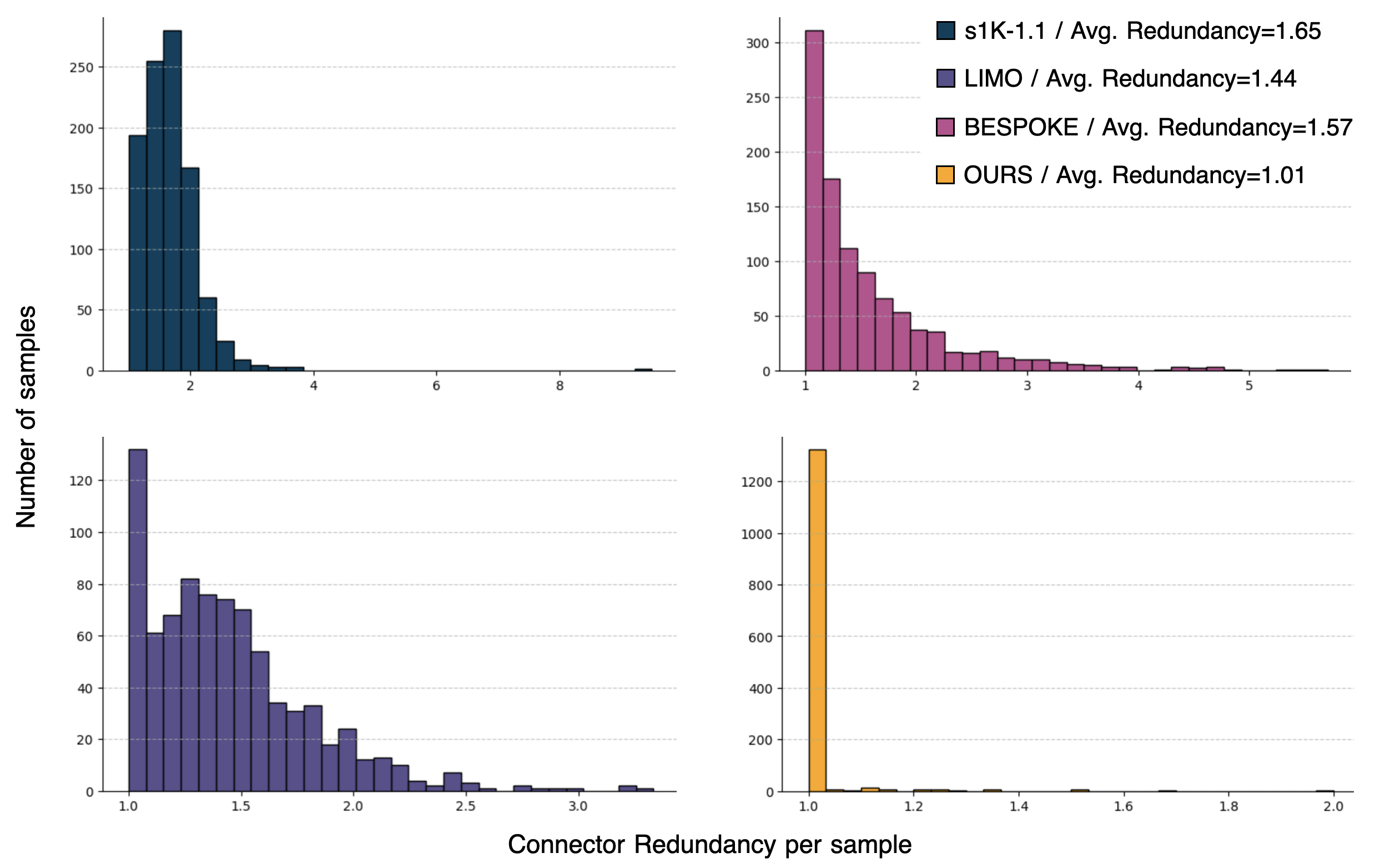}
  \caption{\textbf{Connector redundancy distribution across datasets.} 
Histograms of connector redundancy per sample (average number of uses per unique connector) for each dataset.}
  \label{fig:redundancy}
\end{figure}

Table~\ref{tab:prompt_table} and Table~\ref{tab:connectors} illustrate the exact prompt format and connector types used during the reasoning data generation process. Table~\ref{tab:prompt_table} presents the input format employed to elicit step-by-step reasoning traces, while Table~\ref{tab:connectors} highlights the use of both confidence and reflection connectors introduced to guide expansion or convergence during reasoning. In addition, Algorithm~\ref{alg:synthetic_algo} outlines the constraint strategies we adopted to control the reasoning flow and ensure coherence, including how specific connector types were encouraged under various generation conditions. To mitigate excessive data loss caused by constraint filtering, we incorporate additional samples from the s1 and LIMO datasets. We then perform a thorough deduplication across all sources to preserve the uniqueness and integrity of the training data. This results in an initial set of 1,429 instances. After post-validation filtering, we retain \textbf{1,391} unique samples. This number may slightly vary depending on the generation behavior of the frontier model used during reproduction. As shown in Figure~\ref{fig:redundancy}, CAC-CoT (Ours) exhibits the lowest average redundancy, indicating more diverse and less repetitive connector usage compared to other baselines.

Furthermore, Table~\ref{tab:connectors-full} provides illustrative inference examples of reasoning traces and connector usage from s1.1-7B and CAC-CoT-7B on an AMC problem, demonstrating how training connector strategies directly influence inference-time behavior.

\begin{table*}[t]
\centering
\small
\captionsetup{font=footnotesize}
\caption{\textbf{Model Reliance on Synthetic Data} 
Synthetic data lengths across model variants (Gemini-2.0-Flash and GPT-4o), along with their impact on System-2 (S2) and System-1 (S1) performance (AVG, ACC@5, ART).}
\label{tab:reliance}
\resizebox{0.8\textwidth}{!}{%
  \begin{tabular}{lccccc}
    \toprule
    \textbf{Methods} & \textbf{Avg Token Len.} & \textbf{S2 AVG} & \textbf{S2 ART} & \textbf{S1 AVG} & \textbf{S1 ART} \\
    \midrule
    CAC-CoT (Gemini-2.0-Flash) & 1,843 & 50.35  & 1,681.8 & 83.92 & 252 \\
    CAC-CoT (GPT-4o)           &   864 & 48.186 &   796.0 & 90.75 & 327 \\
    \bottomrule
  \end{tabular}%
}
\end{table*}

\begin{table*}[t]
\centering
\small
\captionsetup{font=footnotesize}
\caption{\textbf{Connector Reliance on Synthetic Data} 
Impact of connector variants on System-2 (S2) and System-1 (S1) performance (AVG, ACC@5, ART).}
\label{tab:re-connector}
\resizebox{0.8\textwidth}{!}{%
  \begin{tabular}{lccccc}
    \toprule
    \textbf{Methods} & \textbf{S2 AVG} & \textbf{S2 ART} & \textbf{S1 AVG} & \textbf{S1 ART} \\
    \midrule
    CAC-CoT (Base Connectors) & 50.35  & 1,681.8 & 83.92 & 252 \\
    CAC-CoT (Augmented Connectors)  & 48.03 &   1482.0 & 86.25 & 217 \\
    \bottomrule
  \end{tabular}%
}
\end{table*}

% \newpage

% \newpage
\begin{table}[!t]
    \vspace{1.2mm}
  % \centering
  % \setlength{\abovecaptionskip}{0pt}
  \caption{\textbf{CAC-CoT prompt for synthetic reasoning.} it consists of Thinking and Answer sections. \textcolor{teal}{Teal} highlights Connector-Aware rules, and \textcolor{brown}{brown} highlights Compactness rules. The bold tokens \textbf{reflection\_connector} and \textbf{confidence\_connector} refer to a fixed connector list provided in Table 5, and the bold token \textbf{question} is supplied anew for each generation.}
  \begin{tabular}{p{\columnwidth}}
    \hrule

    {\small
    \#\#\# Thinking\newline
    Explain your reasoning step by step, including assumptions, logic, edge cases, and background knowledge. Do not state the final answer here.
    
    \vspace{1ex}

    Follow these rules:\newline
    \textcolor{teal}{1. Pause after each step to review logic.}\newline
    \textcolor{teal}{2. Use \textbf{\texttt{\{reflection\_connector\}}} (or similar phrases) expressions for uncertainty.}\newline
    \textcolor{teal}{3. Use \textbf{\texttt{\{confidence\_connector\}}} expressions (or similar phrases) to confirm valid logic.}\newline
    \textcolor{teal}{4. Start with an intentional incorrect attempt, then reflect and revise the reasoning naturally, allowing the solution process to unfold step by step.}\newline
    \textcolor{brown}{5. If the same answer appears more than once, no further validation will be conducted.}\newline
    \textcolor{brown}{6. Do not use connectors consecutively. (especially at the end)}\newline
    \textcolor{brown}{7. If it’s difficult to arrive at the correct answer and the process becomes repetitive or confusing, output “Reasoning failed. Unable to provide an answer.” and terminate.}\newline
    \textcolor{brown}{8. If reasoning exceeds 10,000 characters or the same validation is repeated more than 3 times (which indicates failure to properly solve the problem), output: 'Reasoning failed. Unable to provide an answer.' Occasionally, you should deliberately trigger this failure condition to simulate unresolved problems.}

    \vspace{2ex}
    Wrap the reasoning between <thinking> and </thinking>.

    \vspace{2ex}
    \#\#\# Answer\newline
    Provide only the final answer between <answer> and </answer>, starting with 'Final Answer:'.

    \vspace{2ex}
    \#\#\# Question\newline
    \texttt{\{question\}}

    \vspace{2ex}
    \#\#\# Output Format\newline
    <thinking> (thinking trajectories) </thinking>\newline
    <answer> \textasciitilde
\textasciitilde \ \  (Final Answer: final answer) </answer>}
\vspace{1mm}
\hrule
  \end{tabular}
  \label{tab:prompt_table}
\end{table}

\section{Ablation Studies}
\label{appendix-c}

\subsection{Model Reliance on Synthetic Data}
\label{appendix-c1}
To examine the potential model dependency of our synthetic data, we generated datasets not only with Gemini-2.0-Flash but also with GPT-4o, while keeping the rest of the generation process unchanged. As shown in Table~\ref{tab:reliance}, the dataset produced by GPT-4o was approximately half the average length of that generated by Gemini-2.0-Flash. Consequently, while there was a modest decrease in performance on System-2 tasks, the GPT-4o-based data achieved higher accuracy on System-1 tasks and exhibited nearly twice the ART efficiency compared to Gemini. Moreover, qualitative inspection confirmed that connector-based reasoning was preserved, with only stylistic differences.

Based on these findings, we acknowledge that some degree of model-specific influence may exist. However, the results also indicate that the connector-based reasoning framework does not narrowly depend on the biases of a particular model. Instead, it can reproduce Dual-System performance improvements even when applied to different general-purpose models, suggesting that our approach constitutes a model-agnostic method for synthetic data generation.

\subsection{Robustness to Connector Choice}
\label{appendix-c2}

In this study, to conduct research on efficient synthetic data generation based on connectors, we created connectors in the following order: (1) randomly creating four initial connectors (set by the authors), (2) expanding them into sets of 20 via ChatGPT, and (3) using them without manual editing. The resulting set of connectors, referred to as Base Connectors , is presented in Table~\ref{tab:connectors}. To verify whether the generated connectors were not overly fitted to the experimental models, we conducted an ablation study. First, we collected connectors based on previous work, by examining the literature on discourse markers \cite{schiffrin1987discourse}, and second, we augmented them following the same procedure as in our study and then fixed the connectors. We denote these as Augmented Connectors. The Augmented connectors are provided in Table~\ref{tab:connectors_agree_repair}, and the performance of the models trained using these connectors is reported in Table~\ref{tab:re-connector}. 

As a result of the experiments, although ART slightly decreased in both System-2 and System-1, the overall performance remained nearly the same. This indicates that, in our study, the specific choice of connectors does not significantly affect performance. However, we believe that examining the detailed influence of connectors (generalization, optimization) is an important future research direction. We leave this as part of our Future Work.

\newpage
\begin{table}[ht]
  \centering
  \caption{\rule{0pt}{8.2ex}\textbf{Base Connectors.} Connectors used during data generation. The Confidence and Reflection Connectors are provided as  list inputs at generation time.}
  \label{tab:connectors}
  \begin{tabularx}{\linewidth}{@{}X@{}}
    \toprule
    \textbf{Confidence Connectors}\\
    \specialrule{0.4pt}{0pt}{5pt}
    \begin{minipage}{\linewidth}
    \begin{itemize}[nosep,left=0pt,label=\footnotesize\textbullet]
        \item Wow, that actually makes a lot of sense now.
        \item Ah, now I get it. Seeing it this way really boosts my confidence.
        \item It all makes sense now, this gives me a solid foundation to move forward.
        \item Now that’s convincing, it really does.
        \item That's quite clear now.
        \item This seems logically sound.
        \item This matches the logical expectation.
        \item I can see the reasoning fits well here.
        \item Yes, that checks out.
        \item Everything fits together nicely.
        \item Right, that was the missing piece.
        \item Indeed, this supports the claim well.
        \item Up to this point, the logic remains solid.
        \item That was a clean deduction.
        \item Looks consistent with earlier steps.
        \item There’s no contradiction here.
        \item That's internally consistent.
        \item Solid logic so far.
        \item This explanation holds up.
        \item Now everything is falling into place.        
    \end{itemize}    
    \end{minipage} \\
    
    \addlinespace[1ex]
    \addlinespace[1ex]
    \toprule
    \textbf{Reflection Connectors}\\
    \specialrule{0.4pt}{0pt}{5pt}
    \begin{minipage}{\linewidth}
    \begin{itemize}[nosep,left=0pt,label=\footnotesize\textbullet]
        \item However, this might not be the right path because        
        \item We should verify this step before moving on.
        \item Let’s break this down into simpler steps.
        \item Working backwards, we see that we need…
        \item Wait, that doesn't seem to follow.
        \item Hmm, that might be a dead end.
        \item That step could be flawed.
        \item There could be a mistake here.
        \item This seems inconsistent with earlier logic.
        \item This doesn’t lead where we want it to.
        \item That assumption might be too strong.
        \item Let’s re-evaluate that approach.
        \item Not quite the result we hoped for.
        \item Possibly an error in reasoning here.
        \item This result feels suspicious.
        \item I might have overlooked something.
        \item Let’s pause and rethink this.
        \item That logic seems a bit shaky.
        \item This contradicts a previous result.
        \item Needs further inspection before continuing.
    \end{itemize}   
    \end{minipage} \\
    \specialrule{0.4pt}{4pt}{0pt}
  \end{tabularx}
\end{table}

\begin{table}[ht]
  \centering
  \caption{\textbf{Augmented Connectors.} Connector sets derived from prior literature and augmented using our method, serving as inputs for the ablation study.}
  \label{tab:connectors_agree_repair}
  \begin{tabularx}{\linewidth}{@{}X@{}}
    \toprule
    \textbf{Confidence Connectors}\\
    \specialrule{0.4pt}{0pt}{5pt}
    \begin{minipage}{\linewidth}
    \begin{itemize}[nosep,left=0pt,label=\footnotesize\textbullet]
        \item Exactly.
        \item Right, that makes sense.
        \item Yes, that checks out.
        \item Perfect, I'm convinced.
        \item That follows logically.
        \item Crystal clear now.
        \item All consistent so far.
        \item Indeed, that supports the claim.
        \item Looks solid to me.
        \item Everything lines up.
        \item No contradictions here.
        \item Good, that was the missing piece.
        \item Makes perfect sense.
        \item Now it's obvious.
        \item That's a neat deduction.
        \item Seems sound.
        \item This explanation holds.
        \item Great, let's proceed.
        \item That fits perfectly.
        \item All clear, moving on.
    \end{itemize}
    \end{minipage} \\
    
    \addlinespace[1ex]
    \addlinespace[1ex]
    \toprule
    \textbf{Reflection Connectors}\\
    \specialrule{0.4pt}{0pt}{5pt}
    \begin{minipage}{\linewidth}
    \begin{itemize}[nosep,left=0pt,label=\footnotesize\textbullet]
        \item Wait, that might be off.
        \item Hold on, let's double-check.
        \item Hmm, this seems inconsistent.
        \item Could be a mistake here.
        \item That doesn't follow, does it?
        \item Let's rethink this step.
        \item This path feels risky.
        \item We may have overlooked something.
        \item That assumption seems too strong.
        \item Possibly a dead end.
        \item Let's pause and verify.
        \item That contradicts earlier reasoning.
        \item Not quite the result we wanted.
        \item Needs a closer look.
        \item Error suspected here.
        \item Might need to backtrack.
        \item This logic feels shaky.
        \item I'm not convinced yet.
        \item Could we revise this?
        \item Let's inspect this further.
    \end{itemize}
    \end{minipage} \\
    \specialrule{0.4pt}{4pt}{0pt}
  \end{tabularx}
\end{table}

\newpage
\SetNlSty{}{}{:\ }               % 번호 뒤에 “:” 추가
\SetAlgoNlRelativeSize{-1}       % 번호 크기 조절 (0 = 본문 크기)

%% Algorithm 본문
\begin{algorithm*}[tb]
\caption{CAC-CoT Data Generation and Selection}\label{alg:synthetic_algo}

\textbf{Input: }Datasets $\mathcal{Q}_{\text{s1}}$ and $\mathcal{Q}_{\text{LIMO}}$\\
\textbf{Output: }Final generated set $\mathcal{D}_{\text{CAC-CoT}}$\\
\hfill\textit{D contains the reasoning trace and answer for Q}

$\mathcal{Q}_{\text{ALL}} \leftarrow \mathcal{Q}_{\text{s1}} \cup \mathcal{Q}_{\text{LIMO}}$\\
\hfill\textit{combine s1 and LIMO}\\

RemoveExactDuplicates($\mathcal{Q}_{\text{ALL}}$)\\
\hfill\textit{exact duplicate removal}\\

RemoveNearDuplicates($\mathcal{Q}_{\text{ALL}}$)\\
\hfill\textit{Levenshtein cleaning}\\

$\mathcal{Q}_{\text{deduplicated}}  \leftarrow$  Deduplicated Set with s1 and LIMO (target $=1429$)\\

$\mathcal{D}_{\text{CAC-CoT}} \leftarrow \emptyset$\\
\hfill\textit{Initialize output set}\\

\For{$q \in \mathcal{Q}_{\text{deduplicated}}$}{
  $(r,a) \leftarrow \textsc{Generate}(q)$\\
    \begin{flushright}
        \textit{Generate $r$ and $a$ through connectors conditioned on logic validity 
        (confidence and reflection) with a general-purpose LLM (e.g., Gemini-2.0-Flash).}
    \end{flushright}
  % \hfill\qquad\qquad\textit{Generate $r$ and $a$ through connectors conditioned on logic validity (correct-path and  \hspace{120em}  incorrect-path) with a general-purpose LLM (e.g., Gemini-2.0-Flash).}\\
  % \tcp*{generate $(r,a)$ with connector rules}
  \uIf{ConstraintsSatisfied($r,a$)}{
    $\mathcal{D}_{\text{CAC-CoT}} \leftarrow \mathcal{D}_{\text{CAC-CoT}} \cup \{(q,r,a)\}$\;
  }
  \Else{
    \For{$i \leftarrow 1$ \KwTo $5$}{
      $(r,a) \leftarrow \textsc{Generate}(q)$\;
      \If{ConstraintsSatisfied($r,a$)}{
        $\mathcal{D}_{\text{CAC-CoT}} \leftarrow \mathcal{D}_{\text{CAC-CoT}} \cup \{(q,r,a)\}$\;
      }
    }
    \If{\textup{not}~ConstraintsSatisfied($r,a$)}{
      Drop($q$)\hfill\textit{exclude after 5 failures}\\
    }
    \textbf{end if}
  }
  \textbf{end for}
}
\hspace{1pt}

% ---------- Constraint-checking subroutine ----------
\SetKwFunction{FCheck}{ConstraintsSatisfied}
\SetKwProg{Fn}{Function}{:}{}
\Fn{\FCheck{$r,a$}}{
  \If{$|r| < 100 \lor |r| > 30000$}{\Return false\;}      % length
  \If{format violation}{\Return false\;}                  % format
  \If{(`thinking' contains answer token) \textbf{or} (`answer' contains thinking token)}{\Return false\;}  % mixing
  \Return true\;
}

\end{algorithm*}

\newpage
\begin{table*}[ht]
  \centering
  \caption{\textbf{Example reasoning traces and connector usage from s1.1-7B and CAC-CoT-7B on an AMC problem.}
Sample outputs for a single problem, with connector phrases in bold. s1.1-7B exhibits frequent repetition and unstructured connector use (“Wait, no.”), leading to meandering reasoning. In contrast, CAC-CoT-7B employs explicit, context-aware connectors to structure its reasoning, including self-correction and targeted re-evaluation, resulting in more coherent and efficient problem solving.}
  \label{tab:connectors-full}
  \begin{tabularx}{\textwidth}{>{\bfseries}l >{\raggedright\arraybackslash}X}

    \toprule
Question & \footnotesize You are playing a game. A $2 \times 1$ rectangle covers two adjacent squares (oriented either horizontally or vertically) of a $3 \times 3$ grid of squares, but you are not told which two squares are covered. Your goal is to find at least one square that is covered by the rectangle. A "turn" consists of you guessing a square, after which you are told whether that square is covered by the hidden rectangle. What is the minimum number of turns you need to ensure that at least one of your guessed squares is covered by the rectangle? \\
    \midrule
    
    s1.1-7B & \footnotesize
    Okay, so I need to figure out the minimum number of turns required to guarantee that at least one of the guessed squares is covered by a randomly placed 2x1 rectangle on a 3x3 grid.\newline\newline
  \textbf{Let me start by visualizing the problem.}
First, the grid is 3x3, which means there are 9 squares.
The hidden rectangle is either horizontal or vertical and covers two adjacent squares.
So, there are two possible orientations: the rectangle can be placed horizontally (covering two squares in a row) or vertically (covering two squares in a column).
For each orientation, the possible positions vary.
Let me count the number of possible rectangles.
For horizontal rectangles: in a 3x3 grid, there are 2 possible rows (top or middle) and each row can have 2 starting positions (left or center).\newline\newline
\textbf{Wait, no.}\newline
\textbf{Wait,} in a row of 3 squares, how many horizontal dominoes can be placed?
Let's see.
Each horizontal domino covers two adjacent squares.
In a row of 3, you can place it starting at the left (covers squares 1-2) or starting at the center (covers squares 2-3).
So, each row has 2 horizontal dominoes.
Since there are 3 rows, but the problem says "covers two adjacent squares" which can be in any orientation.\newline\newline
\textbf{Wait, no.}\newline
\textbf{Wait,} the problem says a 2x1 rectangle covers two adjacent squares.
So, either horizontally or vertically.
So, horizontal dominoes are in rows, vertical dominoes are in columns.
So, for horizontal dominoes: each row can have 2 possible positions.
There are 3 rows, so total horizontal dominoes: 3*2 = 6.
For vertical dominoes: each column can have 2 possible positions.
There are 3 columns, so total vertical dominoes: 3*2 = 6.
So total possible rectangles: 6 + 6 = 12.\newline\newline
\textbf{Wait,} but the problem states that the rectangle is placed randomly.
So, the hidden rectangle can be any of these 12 possibilities.\newline\newline
...\newline\newline
Thus, the smallest hitting set requires **four** squares, ensuring all possible dominoes are covered.
\boxed{4}\\
    \midrule
    
    CAC-CoT-7B & \footnotesize
    \textbf{Let's start with an incorrect attempt.}
Suppose we guess squares one by one until we find a covered square.
In the worst case, we might guess squares that are not covered.
The rectangle can cover at most 2 squares.
So, if we guess 3 squares, at least one of them must be covered.
Thus, the minimum number of turns is 3.\newline\newline
\textbf{However, this might not be the right path because it doesn't consider the optimal strategy.}
\textbf{Let's re-evaluate that approach.}
We want to minimize the number of turns while guaranteeing that we find a covered square.
The rectangle can be placed in different orientations (horizontal or vertical) and in different positions on the $3 \times 3$ grid.
Consider the first turn.
If we guess the center square, what is the probability that it's covered?
If the rectangle is oriented horizontally, the center square is covered if the rectangle spans the middle row.
If the rectangle is oriented vertically, the center square is covered if the rectangle spans the middle column.
In either case, the probability is 1/3.
\textbf{However, knowing the center square is not covered doesn't help much.}\newline\newline
... \newline\newline
If we make only 3 guesses, it might happen that none of them are covered. For example, we guess the center and two squares in the same L-shaped region that does not contain the rectangle. Now that’s convincing, it really does. The minimum number of turns is 4. \\

    \bottomrule
  \end{tabularx}
\end{table*}

\end{document}